\def\year{2018}\relax
\documentclass[letterpaper]{article} 
\usepackage{aaai18}  
\usepackage{times}  
\usepackage{helvet}  
\usepackage{courier}  
\usepackage{url}  
\usepackage{graphicx}  
\usepackage{amsmath}
\usepackage{amsfonts}
\usepackage{multirow}

\begin{document}
%
\title{Improving Review Representations with User Attention \\and Product Attention for Sentiment Classification}

\author{Zhen Wu, Xin-Yu Dai\protect\thanks{Corresponding author.}, Cunyan Yin, Shujian Huang, Jiajun Chen\\
National Key Laboratory for Novel Software Technology, Nanjing University, Nanjing, 210023, China\\
 Collaborative Innovation Center of Novel Software Technology and Industrialization, Nanjing, 210023, China\\
wuz@nlp.nju.edu.cn, \{daixinyu,yincy,huangsj,chenjj\}@nju.edu.cn\\
}
\maketitle
\begin{abstract}
Neural network methods have achieved great success in reviews sentiment classification. Recently, some works achieved improvement by incorporating user and product information to generate a review representation. However, in reviews, we observe that some words or sentences show strong user's preference, and some others tend to indicate product's characteristic. The two kinds of information play different roles in determining the sentiment label of a review. Therefore, it is not reasonable to encode user and product information together into one representation. In this paper, we propose a novel framework to encode user and product information. Firstly, we apply two individual hierarchical neural networks to generate two representations, with user attention or with product attention. Then, we design a combined strategy to make full use of the two representations for training and final prediction. The experimental results show that our model obviously outperforms other state-of-the-art methods on IMDB and Yelp datasets. Through the visualization of attention over words related to user or product, we validate our observation mentioned above.
\end{abstract}

\section{Introduction}
\noindent Sentiment analysis aims to determine people's attitudes towards some topic or the overall polarity to a document, interaction, or event. In recent years, sentiment analysis draws increasing attention of researchers and industries because of the rapid growth of online review sites such as Amazon, Yelp and IMDB. In this work, we focus on the task of document-level review sentiment classification, which is a fundamental task in the field of sentiment analysis and opinion mining~\cite{INR-011}. The task aims to infer the overall sentiment intensity (e.g. 1-5 stars on the review site Yelp) of review documents written by users for products.

Dominating studies follow ~\cite{pang2005seeing,pang2002thumbs} and take sentiment classification as a special case of text classification problem. They usually regard user-marked sentiment polarities or ratings as labels and use machine learning algorithms to build sentiment classifiers with text features. Following the idea, many works devote to designing effective features from text~\cite{pang2002thumbs,qu2010bag} or additional sentiment lexicons~\cite{ding2008holistic,taboada2011lexicon,kiritchenko2014sentiment}. Motivated by the great success of deep learning in computer vision~\cite{krizhevsky2012imagenet}, speech recognition~\cite{dahl2012context} and natural language processing~\cite{bengio2003neural}, more recent methods use neural networks to learn low-dimensional and continuous text representations without any feature engineering~\cite{glorot2011domain,socher2011semi,socher2012semantic,socher2013recursive,kim2014convolutional}. These models achieve very competitive performances in sentiment classification.

Despite neural network based approaches have been quite effective for sentiment classification~\cite{johnson2014effective,tang2015document}, they typically only focus on the text content while ignoring the crucial influences of users and products. It is a common sense that the user's preference and product's characteristic make a significant effect on the ratings. For different users, same word might express different emotional intensity. For example, a lenient user may use ``good" to evaluate an ordinary product while a critical user might use ``good" to express an excellent attitude. Similarly, product's characteristic also have an effect on review ratings. Reviews of high-quality products tend to receive higher ratings compared to those of low-quality products.

In order to incorporate user and product information into sentiment classification,~\citeauthor{tang2015learning}~\shortcite{tang2015learning} introduce a word-level preference matrix and a representation vector for each user and product into CNN sentiment classifier. The model achieves some improvements, but suffers from high model complexity and only considers word-level user and product information rather than semantic-level.~\citeauthor{chen2016neural}~\shortcite{chen2016neural} consider user and product information together and incorporate them into one review representation via attention mechanism~\cite{bahdanau2014neural}. However, in reviews, we observe that some words or sentences show strong user's preference, and some others tend to indicate product's characteristic. For example, for the review ``\textit{the bar area is definitely good `people watching' and i love the modern contemporary d{\'e}cor.}", the word ``\textit{good}", ``\textit{modern}" and ``\textit{contemporary}" describe the characteristic of product, and the ``\textit{love}" shows strong user's sentiment. Opinions are more related to products and emotions are more centered on the user. They are called rational evaluation and emotional evaluation respectively by~\citeauthor{liu2012sentiment}~\shortcite{liu2012sentiment}. Obviously, the two kinds of information have different effects on inferring sentiment label of the review. Intuitively, a review has different latent semantic representations with different views from users or products. Therefore, it is unreasonable to encode user and product information together into one representation.

In this paper, we address the above issues by encoding user and product information, respectively, into two hierarchical networks to generate two individual text representations with user attention or product attention. Then, we design a combined strategy to make most use of the two representations for training and final prediction, which proves effective. The experimental results show that our model obviously outperforms other state-of-the-art methods on IMDB and Yelp datasets. We open the source code in GitHub.\footnote{https://github.com/wuzhen247/HUAPA}

The main contributions of our work are as follows:
\begin{itemize}
\item We propose a novel framework to encode review from two views for sentiment classification. With user attention and product attention, respectively, two representations are generated, which are concatenated for further classification.
\item For better learning the neural network, we introduce a combined strategy to improve review representations. With a weighted loss function, better representations from two views can be achieved and further help sentiment classification.
\item The experimental results demonstrate that our model achieves obvious and consistent improvements compared to all state-of-the-art methods. Some visualization cases also validate the effectiveness and interpretability of our method.
\end{itemize}

\section{Background}
\subsection{Long Short-Term Memory}
Long Short-Term Memory(LSTM)~\cite{Hochreiter1997Long} is widely used for text modeling because of its excellent performance on sequence modeling, especially for long documents. In order to address the problem of long-term dependencies, the LSTM architecture introduces a memory cell that is able to preserve cell state over long periods of time.

There are three gates to protect and control the state flow in LSTM unit. At each time step $t$, given an input vector $\mathbf{x}_{t}$, the current cell state $\mathbf{c}_{t}$ and hidden state $\mathbf{h}_{t}$ can be updated with previous cell state $\mathbf{c}_{t-1}$ and hidden state $\mathbf{h}_{t-1}$ as follows:
\begin{align}
  {\begin{bmatrix}\mathbf{i}_{t}\\ \mathbf{f}_{t}\\ \mathbf{o}_{t}\end{bmatrix}} &= {\begin{bmatrix}\sigma\\ \sigma\\ \sigma\end{bmatrix}} \left ( \mathbf{W} \left [ \mathbf{h}_{t-1}; \mathbf{x}_{t} \right ]+\mathbf{b} \right ), \\
  \mathbf{\hat c}_{t} &= \tanh \left ( \mathbf{W}_{c} \left [ \mathbf{h}_{t-1}; \mathbf{x}_{t} \right ]+\mathbf{b}_{c} \right ), \\
  \mathbf{c}_{t} &= \mathbf{f}_{t}\odot \mathbf{\hat c}_{t-1} + \mathbf{i}_{t}\odot \mathbf{\hat c}_{t}, \\
  \mathbf{h}_{t} &= \mathbf{o}_{t}\odot \tanh(\mathbf{c}_{t}),
\end{align}
where $\mathbf{i}_{t}$, $\mathbf{f}_{t}$ and $\mathbf{o}_{t}$ are gate activations and in $\left [0,1\right ]$, $\sigma$ is the logistic sigmoid function and $\odot$ stands for element-wise multiplication. Intuitively, the forget gate $\mathbf{f}_{t}$ controls the extent to which the previous memory cell is forgotten, the input gate $\mathbf{i}_{t}$ controls how much each unit is updated, and the output gate $\mathbf{o}_{t}$ controls the exposure of the internal memory state. The hidden state $\mathbf{h}_{t}$ denotes output information of LSTM unit's internal memory cell.

In order to increase the amount of input information available to the network, a more common approach is to adopt bidirectional LSTM to model text semantics both from forward and backward. For sequence vectors $\left [ \mathbf{x}_{1}, \mathbf{x}_{2},\cdots,\mathbf{x}_{T} \right ]$, the forward LSTM reads sequence from $\mathbf{x}_{1}$ to $\mathbf{x}_{T}$ and the backward LSTM reads sequence from $\mathbf{x}_{T}$ to $\mathbf{x}_{1}$. Then we concatenate the forward hidden state $\overrightarrow{\mathbf{h}_{t}}$ and backward hidden state $\overleftarrow{\mathbf{h}_{t}}$, i.e., $\mathbf{h}_{t}=\left [\overrightarrow{\mathbf{h}_{t}}; \overleftarrow{\mathbf{h}_{t}} \right ]$, where the $\left [\cdot;\cdot \right ]$ denotes concatenation operation. The $\mathbf{h}_{t}$ summarizes the information of the whole sequence centered around $\mathbf{x}_{t}$.

\subsection{Attention Mechanism}
Inspired by human visual attention, the attention mechanism is proposed by~\citeauthor{bahdanau2014neural}~\shortcite{bahdanau2014neural} in machine translation, which is introduced into the Encoder-Decoder framework to select the reference words in source language for words in target language. It is also used in image caption generation~\cite{xu2015show}, parsing~\cite{vinyals2015grammar}, natural language question answering~\cite{sukhbaatar2015end}.~\citeauthor{yang2016hierarchical}~\shortcite{yang2016hierarchical} and ~\citeauthor{chen2016neural}~\shortcite{chen2016neural} explore hierarchical attention mechanism to select informative words or sentences for the semantics of document.

\subsection{Document-level Sentiment Classification}
Document-level sentiment classification usually aims to predict the corresponding sentiment label of review text. In general, a review is written by a user $u\in U$ for a product $p\in P$. We denote the review as a document $d$ with $n$ sentences $\left \{ s_{1},s_{2},\cdots,s_{n} \right \}$ and $l_{i}$ is the length of $i$-th sentence. The $i$-th sentence $s_{i}$ consists of $l_{i}$ words $\{w_{i1},w_{i2},\cdots,w_{il_{i}}\}$. For modelling document-level text semantics, we can employ bidirectional LSTM with hierarchical structure to obtain document representation. In word level, each word $w_{ij}$ is mapped to its embedding $\mathbf{w}_{ij}\in \mathbb{R}^d$. BiLSTM network receives $\left [ \mathbf{w}_{i1},\mathbf{w}_{i2},\cdots,\mathbf{w}_{il_{i}} \right ]$ and generates hidden states $\left [ \mathbf{h}_{i1},\mathbf{h}_{i2},\cdots,\mathbf{h}_{il_{i}} \right ]$. Then we can fetch the last hidden state $\mathbf{h}_{il_{i}}$ or feed $\left [ \mathbf{h}_{i1},\mathbf{h}_{i2},\cdots,\mathbf{h}_{il_{i}} \right ]$ to an average pooling layer or use attention mechanism to obtain the sentence representation $\mathbf{s}_{i}$. In sentence level, we feed the generated sentence representations $\left [ \mathbf{s}_{1},\mathbf{s}_{2},\cdots,\mathbf{s}_{n} \right ]$ into BiLSTM and then obtain the document representation $\mathbf{d}$ in a similar way. Finally, $\mathbf{d}$ is taken as feature of softmax classifier to predict sentiment label of the review.

\begin{figure*}[t]
\centering
\includegraphics[width=1.0\textwidth]{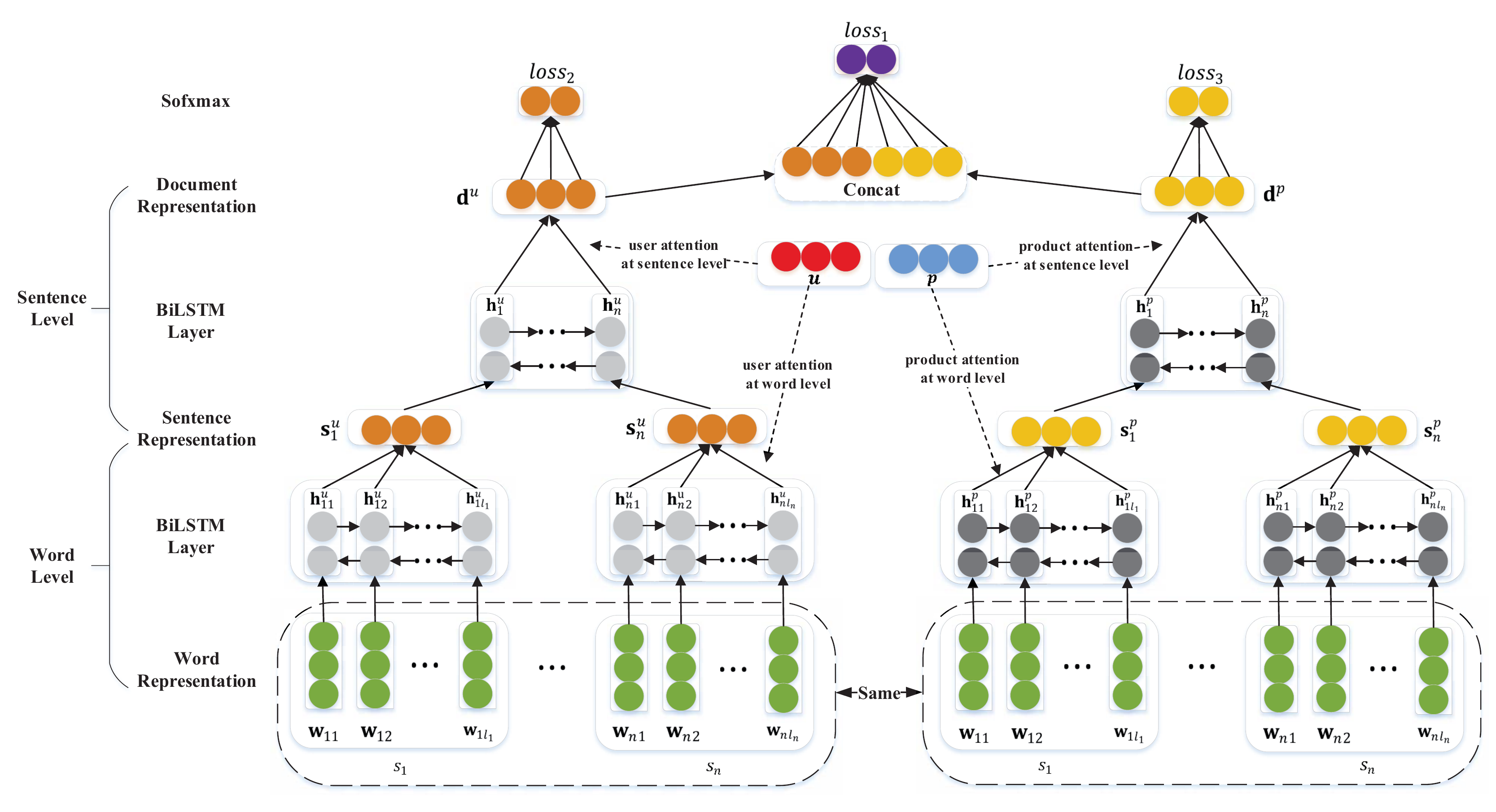}
\caption{The architecture of Hierarchical User Attention and Product Attention neural network.}
\label{fig:HUAPA}
\end{figure*}

\section{Methods}
It is obvious that not all words/sentences contribute equally to the review text semantics for different users and different products. In addition, in reviews, we observe that some words or sentences show strong user's preference, and some others tend to indicate product's characteristic. The two kinds of information play different roles in inferring the sentiment label of reviews, which implies a review has different latent semantic representations in user's view and product's view. In order to address the issue, we propose a novel framework to incorporate user and product information into sentiment classification.

\subsection{Framework}
 We refer to our model as HUAPA for convenience. An illustration of HUAPA is shown in Figure \ref{fig:HUAPA}. It consists of two components mainly: a hierarchical user attention network and a hierarchical product attention network. With attention mechanism, the former incorporates user information into review document modeling while the latter incorporates product information. Then, we concatenate the two review representations as the final representation to predict user's overall sentiment on a review about a product. In addition, we design a combined strategy to enhance review representations for sentiment classification. Specifically,  we add a softmax classifier respectively to three review representation $\mathbf{d}^{u}$,  $\mathbf{d}^{p}$ and $\mathbf{d}$, and introduce a weighted loss function as optimization objective, which proves effective. We will present the details of HUAPA in the following sections.

\subsection{Hierarchical User Attention}
\begin{table*}[!htp]
\centering
\begin{tabular}{c|c|c|c|c|c|c|c|c}
\hline
Datasets  & \#classes & \#docs & \#users & \#products & \#docs/user & \#docs/product & \#sens/doc & \#words/sen  \\
\hline
IMDB  & 10 & 84,919 & 1,310 & 1,635 & 64.82 & 51.94 & 16.08 & 24.54 \\
Yelp 2013  & 5 & 78,966 & 1,631 & 1,633 & 48.42 & 48.36 & 10.89 & 17.38 \\
Yelp 2014  & 5 & 231,163 & 4,818 & 4,194 & 47.97 & 55.11 & 11.41 & 17.26 \\
\hline
\end{tabular}
\caption{Statistics of IMDB, Yelp2013 and Yelp2014 datasets}
\label{tab:statistics}
\end{table*}
From the user's point of view, not all words reflect equally user's preference or sentiment. In order to address the issue, we use user attention mechanism to extract user-specific words that are significant to the meaning of sentence. Finally, the sentence representation is aggregated by the representations of those informative words. Formally, the enhanced sentence representation $\mathbf{s}_{i}^{u}$ is a weighted sum of word-level hidden states in user's view as:
\begin{equation}
\mathbf{s}_{i}^{u}=\sum_{j=1}^{l_{i}} \alpha_{ij}^{u} \mathbf{h}_{ij}^{u},
\end{equation}
where $\mathbf{h}_{ij}^{u}$ is the hidden state of the $j$-th word in the $i$-th sentence, $\alpha_{ij}^{u}$ is the attention weight of $\mathbf{h}_{ij}^{u}$ and measures the importance of the $j$-th word for current user. We map each user $u$ into a continuous and real valued vector $\mathbf{u}\in \mathbb{R}^{d_{u}}$, where $d_u$ denotes the dimension of user embeddings. Specifically, the attention weight $\alpha_{ij}^{u}$ for each hidden state can be defined as:
\begin{align}
  e(\mathbf{h}_{ij}^{u}, \mathbf{u}) &= (\mathbf{v}_{w}^{u} )^{\top} \tanh\left (\mathbf{W}_{wh}^{u} \mathbf{h}_{ij}^{u}+\mathbf{W}_{wu}^{u}\mathbf{u} +\mathbf{b}_{w}^{u}  \right ), \\
  \alpha_{ij}^{u} &= \frac{\exp\left(e\left(\mathbf{h}_{ij}^{u}, \mathbf{u} \right ) \right )} {\sum_{k=1}^{l_i}\exp\left(e\left(\mathbf{h}_{ik}^{u}, \mathbf{u} \right ) \right )}
\end{align}
where $\mathbf{v}_{w}^{u}$ is weight vector and $\left (\mathbf{v}_{w}^{u}\right )^{\top}$ represents its transpose, $\mathbf{W}_{wh}^{u}$ and $\mathbf{W}_{wu}^{u}$ are weight matrices, $e(\cdot)$ is a score function which scores the importance of words for composing sentence representation about current user.

Similarly, different sentences contribute unequally to document semantics for users. Therefore, in sentence level, we also use a attention mechanism with user vector $\mathbf{u}$ in word level to generate the document representation. The document representation $\mathbf{d}^{u}$ in user's view is obtained via:
\begin{align}
  e\left(\mathbf{h}_{i}^{u}, \mathbf{u} \right ) &= \left (\mathbf{v}_{s}^{u}\right )^{\top} \tanh\left (\mathbf{W}_{sh}^{u} \mathbf{h}_{i}^{u}+\mathbf{W}_{su}^{u}\mathbf{u} +\mathbf{b}_{s}^{u}  \right ), \\
  \beta_{i}^{u} &= \frac{\exp\left(e\left(\mathbf{h}_{i}^{u}, \mathbf{u} \right ) \right )} {\sum_{k=1}^{n}\exp\left(e\left(\mathbf{h}_{k}^{u}, \mathbf{u} \right ) \right )}, \\
  \mathbf{d}^{u} &= \sum_{i=1}^{n} \beta_{i}^{u} \mathbf{h}_{i}^{u},
\end{align}
where $\mathbf{h}_{i}^{u}$ is the hidden state of the $i$-th sentence in a review document, $\beta_{i}^{u}$ is the weight of hidden state $\mathbf{h}_{i}^{u}$ in sentence level and can be calculated similar to the word level.

\subsection{Hierarchical Product Attention}
It is also true for different products that every word or sentence contributes different information to the text semantics. Based on the common sense, hierarchical product attention incorporates product information into review representation similar to hierarchical user attention. In product's view, the sentence representation $\mathbf{s}_{i}^{p}$ and document representation $\mathbf{d}^{p}$ of a review are obtained formally as:
\begin{align}
\mathbf{s}_{i}^{p}&=\sum_{j=1}^{l_{i}} \alpha_{ij}^{p} \mathbf{h}_{ij}^{p}, \\
\mathbf{d}^{p} &= \sum_{i=1}^{n} \beta_{i}^{p} \mathbf{h}_{i}^{p},
\end{align}
where $\alpha_{ij}^{p}$ and $\beta_{i}^{p}$ are the weight of hidden state $\mathbf{h}_{ij}^{p}$ in word level and $\mathbf{h}_{i}^{p}$ in sentence level respectively.

\subsection{Combined Strategy}
In order to make full use of document representation $\mathbf{d}^{u}$ and $\mathbf{d}^{p}$, we design a combined strategy for training and the final prediction.

Since document representation $\mathbf{d}^{u}$ and $\mathbf{d}^{p}$ are high level representations of review in user's view and product's view respectively. Hence, we concatenate them as the final review representation for sentiment classification without feature engineering:
\begin{equation}
    \mathbf{d} = \left [\mathbf{d}^{u}; \mathbf{d}^{p} \right ].
\end{equation}
Specifically, we use a linear layer and a softmax layer to project review representation $\mathbf{d}$ into review sentiment distribution of $C$ classes:
\begin{equation}
    p = \rm{softmax} \left (\mathbf{W}\mathbf{d}+\mathbf{b} \right ).
\end{equation}
In our model, the cross-entropy error between ground truth distribution of review sentiment and $p$ is defined as $loss_1$:
\begin{equation}
    loss_{1} =- \sum_{d\in T}\sum_{c=1}^{C}p_{c}^{g}(d)\cdot \log\left (p_{c}(d) \right ),
\end{equation}
where $p_{c}^{g}$ is the probability of sentiment label $c$ with ground truth being 1 and others being 0, $T$ represents the training set.

To make the review representations $\mathbf{d}^{u}$ and $\mathbf{d}^{p}$ both have certain predictive capability and futher improve performance for sentiment classification, we add a softmax classifier respectively to $\mathbf{d}^{u}$ and $\mathbf{d}^{p}$. The corresponding losses are defined as follows:
\begin{align}
    p^{u} &= {\rm{softmax}}\left(\mathbf{W}^{u}\mathbf{d}^{u}+\mathbf{b}^{u} \right), \\
    loss_{2} &=- \sum_{d\in T}\sum_{c=1}^{C}p_{c}^{g}(d)\cdot \log\left (p_{c}^{u}(d) \right ), \\
    p^{p} &= {\rm{softmax}}\left (\mathbf{W}^{p}\mathbf{d}^{p}+\mathbf{b}^{p} \right ), \\
    loss_{3} &=- \sum_{d\in T}\sum_{c=1}^{C}p_{c}^{g}(d)\cdot \log\left (p_{c}^{p}(d) \right ),
\end{align}
where $p^{u}$ is the predicted sentiment distribution of with user's view and $p^{p}$ is the predicted sentiment distribution of with product's view. The final loss of our model is a weighted sum of $loss_{1}$, $loss_{2}$ and $loss_{3}$ as:
\begin{equation}
    L=\lambda_{1}loss_{1}+\lambda_{2}loss_{2}+\lambda_{3}loss_{3}.
\end{equation}
The $loss_{2}$ and $loss_{3}$ are introduced as supervised information to improve review representations and further help sentiment classification. Note that, we predict review sentiment label according to the distribution $p$ because it contains both user information and product information.

\begin{table*}[!htbp]
\centering
\begin{tabular}{c||c|c||c|c||c|c}
\hline
\multirow{2}*{Models} & \multicolumn{2}{c||}{IMDB} & \multicolumn{2}{c||}{Yelp 2013} & \multicolumn{2}{c}{Yelp 2014} \\
\cline{2-7}
\multicolumn{1}{c||}{}  & Acc. & RMSE & Acc. & RMSE & Acc. & RMSE\\
\hline
\multicolumn{7}{c}{\emph{Models without user and product information}}\\
\hline
Majority & 0.196 & 2.495 & 0.411 & 1.060 & 0.392 & 1.097 \\
Trigram & 0.399 & 1.783 & 0.569 & 0.814 & 0.577 & 0.804 \\
TextFeature & 0.402 & 1.793 & 0.556 & 0.845 & 0.572 & 0.800 \\
AvgWordvec+SVM & 0.304 & 1.985 & 0.526 & 0.898 & 0.530 & 0.893 \\
SSWE+SVM & 0.312 & 1.973 & 0.549 & 0.849 & 0.557 & 0.851 \\
Paragraph Vector & 0.341 & 1.814 & 0.554 & 0.832 & 0.564 & 0.802\\
RNTN+Recurrent & 0.400 & 1.764 & 0.574 & 0.804 & 0.582 & 0.821 \\
UPNN(CNN and no UP) & 0.405 & 1.629 & 0.577 & 0.812 & 0.585 & 0.808 \\
NSC & 0.443 & 1.465 & 0.627 & 0.701 & 0.637 & 0.686 \\
NSC+LA & 0.487 & 1.381 & 0.631 & 0.706 & 0.630 & 0.715 \\
NSC+LA(BiLSTM) & 0.490 & 1.325 & 0.638 & 0.691 & 0.646 & 0.678 \\
\hline
\multicolumn{7}{c}{\emph{Models with user and product information}}\\
\hline
Trigram+UPF & 0.404 & 1.764 & 0.570 & 0.803 & 0.576 & 0.789 \\
TextFeature+UPF & 0.402 & 1.774 & 0.561 & 1.822 & 0.579 & 0.791 \\
JMARS & N/A & 1.773 & N/A & 0.985 & N/A & 0.999 \\
UPNN(CNN) & 0.435 & 1.602 & 0.596 & 0.784 & 0.608 & 0.764 \\
UPNN(NSC) & 0.471 & 1.443 & 0.631 & 0.702 & N/A & N/A \\
LUPDR & 0.488 & 1.451 & 0.639 & 0.694 & 0.639 & 0.688 \\
NSC+UPA & 0.533 & 1.281 & 0.650 & 0.692 & 0.667 & 0.654 \\
NSC+UPA(BiLSTM) & 0.529 & 1.247 & 0.655 & 0.672 & 0.669 & 0.654 \\
\hline
HUAPA & \textbf{0.550} & \textbf{1.185} & \textbf{0.683} & \textbf{0.628} & \textbf{0.686} & \textbf{0.626} \\
\hline
\end{tabular}
\caption{Reviews sentiment classification results. Acc.(Accuracy, higher is better) and RMSE(lower is better) are the evaluation metrics. The best performances are in bold. HUAPA outperforms previous best state-of-the-art method significantly($p<0.01$).}\label{tab:results}
\end{table*}

\section{Experiments}
We conduct experiments on several real-world datasets to validate the effectiveness of our model and report empirical results in this section.

\subsection{Experiments Settings}
We conduct experiments on three sentiment classification datasets\footnote{http://ir.hit.edu.cn/~dytang/paper/acl2015/dataset.7z} with user and product information, which are from IMDB and Yelp Dataset Challenge in 2013 and 2014~\cite{tang2015learning}. The statistics of the datasets are summarized in Table \ref{tab:statistics}. The datasets are split into three parts, $80\%$ for training, $10\%$ for validation, and the remaining $10\%$ for test. We use standard $Accuracy$ to measure the overall sentiment classification performance, and $RMSE$ to measure the divergences between predicted sentiment label and ground truth label. They are defined as follows:
\begin{align}
   &Accuracy = \frac{T} {N}, \\
   &RMSE = \sqrt{\frac{\sum_{k=1}^{N}\left (gd_k-pr_k \right )^2}{N}},
\end{align}
where $T$ is the numbers of predicted sentiment labels that are same with ground truth sentiment labels of reviews, $N$ is the numbers of review documents, $gd_k$ represents the ground sentiment label, and $pr_k$ denotes predicted sentiment label.

We pre-train the 200-dimensional word embeddings on each dataset with SkipGram~\cite{mikolov2013distributed}. The word embeddings are not fine-tuned when training, so hierarchical user attention and hierarchical product attention use same word embeddings. We set the user embeddings dimension and product embeddings dimension to be 200, and randomly initialize them from a uniform distribution $U(-0.01,0.01)$. The dimensions of hidden states in LSTM cell are set to 100. In this setting, a bidirectional LSTM gives us 200 dimensional output for word/sentence representation. To speed up training, we limit that a review document has 40 sentences at most and every sentence has no more than 50 words.
We use Adam~\cite{kingma2014adam} to update parameters when training and empirically set initial learning rate to be 0.005. Finally, We select the best parameters based on performance on the validation set, and evaluate the parameters on the test set. Note that, we do not use any regularization or dropout~\cite{srivastava2014dropout} to improve performance of the model.

\subsection{Baselines}
\begin{table*}[!htbp]
\centering
\begin{tabular}{c||c|c||c|c||c|c}
\hline
\multirow{2}*{Models} & \multicolumn{2}{c||}{IMDB} & \multicolumn{2}{c||}{Yelp 2013} & \multicolumn{2}{c}{Yelp 2014} \\
\cline{2-7}
\multicolumn{1}{c||}{}  & Acc. & RMSE & Acc. & RMSE & Acc. & RMSE\\
\hline
NSC+LA(BiLSTM) & 0.490 & 1.325 & 0.638 & 0.691 & 0.646 & 0.678 \\
HUA & 0.521 & 1.300 & 0.649 & 0.691 & 0.663 & 0.661 \\
HPA & 0.493 & 1.326 & 0.641 & 0.681 & 0.646 & 0.678 \\
HUAPA & \textbf{0.550} & \textbf{1.185} & \textbf{0.683} & \textbf{0.628} & \textbf{0.686} & \textbf{0.626} \\
\hline
\end{tabular}
\caption{Effect of user attention and product attention. HUA only uses user information and local text, and HPA only uses product information and local text.}\label{tab:usrprdattention}
\end{table*}

\begin{table*}[!htbp]
\centering
\begin{tabular}{c|c|c||c|c||c|c||c|c}
\hline
\multirow{2}*{$\lambda_1$} & \multirow{2}*{$\lambda_2$} & \multirow{2}*{$\lambda_3$} & \multicolumn{2}{c||}{IMDB} & \multicolumn{2}{c||}{Yelp 2013} & \multicolumn{2}{c}{Yelp 2014} \\
\cline{4-9}
 {}& {} & {} & Acc. & RMSE & Acc. & RMSE & Acc. & RMSE\\
\hline
1.0 & 0.0 & 0.0 & 0.538 & 1.229 & 0.669 & 0.658 & 0.675 & 0.647 \\
0.7 & 0.3 & 0.0 & 0.541 & 1.239 & 0.672 & 0.644 & 0.680 & 0.641 \\
0.7 & 0.0 & 0.3 & 0.540 & 1.287 & 0.675 & 0.646 & 0.679 & 0.633 \\
0.4 & 0.3 & 0.3 & \textbf{0.550} & \textbf{1.185} & \textbf{0.683} & \textbf{0.628} & \textbf{0.686} & \textbf{0.626} \\
\hline
\end{tabular}
\caption{Effect of the different weighted loss.}\label{tab:weighteffect}
\end{table*}
We compare our model HUAPA with several baseline methods for document-level review sentiment classification:

\textbf{Majority} assigns the majority sentiment label in training set to each review document in test set.

\textbf{Trigram} uses unigrams, bigrams and trigrams as features to train a SVM classifier with LibLinear~\cite{fan2008liblinear}.

\textbf{TextFeature} extracts sophisticated text features to train SVM classifier, such word/character n-grams, sentiment lexicon features, cluster features, etc.~\cite{kiritchenko2014sentiment}.

\textbf{UPF} extracts user leniency and corresponding product popularity features~\cite{gao2013modeling} from training data, and further concatenates them with the features in \textbf{Trigram} and \textbf{TextFeature}.

\textbf{AvgWordvec} averages word embeddings of a document to generate document representation, then feeds it into a SVM classifier as features.

\textbf{SSWE} learns sentiment-specific word embeddings (SSWE)~\cite{tang-EtAl:2014:P14-1}, and uses max/min/average pooling to obtain document representation which is used as features for a SVM classifier.

\textbf{RNTN + RNN} uses the Recursive Neural Tensor Network (RNTN)~\cite{socher2013recursive} to obtain sentence representations, then feeds them into the Recurrent Neural Network (RNN). Afterwards, the hidden vectors of RNN are averaged to generate document representation for sentiment classification.

\textbf{Paragraph Vector} implements the Distributed Memory Model of Paragraph Vectors~\cite{le2014distributed} for document sentiment classification. The window size is tuned on validation set.

\textbf{JMARS} is a recommendation algorithm~\cite{diao2014jointly}, which uses the information of users and aspects with collaborative filtering and topic modeling to predict document sentiment rating.

\textbf{UPNN} introduces a word-level preference matrix and a representation vector for each user and each product into CNN sentiment classifier~\cite{kim2014convolutional}. The meaning of words can modified in the input layer with the preference matrix. Finally, it concatenates the user/product representation vectors with generated review representation as features fed into softmax layer~\cite{tang2015learning}.

\textbf{LUPDR} uses recurrent neural network to embed temporal relations of reviews into the categories of distributed user and product representations learning for the sentiment classification of reviews~\cite{chen2016learning}.

\textbf{NSC} uses hierarchical LSTM model to encode review text for sentiment classification.

\textbf{NSC+LA} implements the idea of local semantic attention~\cite{yang2016hierarchical} based on \textbf{NSC}.

\textbf{NSC+UPA} puts user and product information account together and uses hierarchical LSTM model with attention mechanism to generate a review representation for sentiment classification.

~\citeauthor{chen2016neural}~\shortcite{chen2016neural} do not implement the model \textbf{NSC+UPA} and \textbf{NSC+LA} with bidirectional LSTM. To make the experimental results more convincing, we implement and train them in our experimental settings. In addition to \textbf{LUPDR} and the models related to \textbf{NSC}, we report the results in~\cite{tang2015learning} since we use the same datasets for other baseline methods above.

\subsection{Model Comparisons}
The experimental results are given in Table \ref{tab:results}, which are divided into two parts: the models only using the local text information, and the models incorporating both local text information and the global user and product information.

From the first part, we can see that the majority performs very poor because it does not use any text, user, and product information. Compared to the methods taking SVM as classifier, hierarchical neural networks achieve better performances generally. In addition, the results show NSC+LA obtains a considerable improvements based on NSC, which proves that importance of selecting more meaningful words and sentences in sentiment classification. It is also a main reason that attention mechanism is introduced into sentiment classification.

From the second part, we observe that the methods considering user and product information achieve more or less improvements compared to the corresponding methods in the first part. For example, TextFeature+UPF achieves 0.5\% improvement and NSC+UPA obtains 2.3\% improvement on Yelp2013 in accuracy. The comparisons indicate that the user and product information is helpful for sentiment classification.

The experimental results show that our proposed model with user attention and product attention achieves best performance on all datasets. We can see improvements regardless of data scale. For smaller dataset such as Yelp2013 and IMDB, our model outperforms the previous best state-of-the-art method by 2.8\% and 1.7\% respectively in accuracy. This finding is consistent on larger dataset. Our model achieves improvement by 1.7\% on dataset Yelp2014 in accuracy. The observations demonstrate that our model incorporates user and product information in a more effective way, which finally improves review representations for sentiment classification.

\subsection{Model Analysis: Effect of User Attention and Product Attention}
To investigate the effects of single user attention or product attention, we also implement independent hierarchical user attention network (HUA) and hierarchical product attention network (HPA). Table \ref{tab:usrprdattention} shows the performance of single attention mechanism with user or product information. From the table, we can observe that:
\begin{itemize}
\item Compared to the model NSC+LA(BiLSTM) only using local semantic attention, HUA and HPA both achieve some improvements, which validates the rationality of incorporating user and product into sentiment classification via attention mechanism. The results also indicate that user attention or product attention can capture more information related to sentiment.
\item The user information is more effective than the product information to enhance review representations. Although some words or sentences in reviews show product's characteristic, the ratings are finally decided by users. Hence, it is reasonable that the discrimination of user's preference is more obvious than product's characteristic.
\item Compared to single user attention or product attention, our model achieves better performance, which indicates that user and product information both contribute to our model. The results demonstrate our user attention and product attention mechanism can catch the specific user's preference and product's characteristic.
\end{itemize}

\subsection{Model Analysis: Effect of the Different Weighted Loss}
The $\lambda_{1}$, $\lambda_{2}$, and $\lambda_{3}$ respectively represent the weight of $loss_1$, $loss_2$ and $loss_3$. We investigate the effect of different weighted loss by empirically adjusting their proportion.  When $\lambda_{2}$ is set to $0$, we do not use $loss_2$ to enhance the review representations. Similarly, we set $\lambda_{3}$ to $0$ to avoid the effect of $loss_3$. The experimental results are in Table \ref{tab:weighteffect}. From the Table \ref{tab:weighteffect} and Table \ref{tab:results}, we can observe that:

\begin{itemize}
\item Compared to other state-of-the-art methods, our model without $loss_2$ and $loss_3$ also achieve consistent improvements on the three datasets. It indicates that our attention mechanism is more effective in incorporating user information and product information.
\item When considering $loss_2$ or $loss_3$, HUAPA both obtains some improvements. It is obvious that full HUAPA model achieves best performance. The results demonstrate that with the designed combined strategy, better review representations from two views can be achieved and further help sentiment classification.
\end{itemize}

\subsection{Case Study for Visualization of Attention}
\begin{figure}[ht]
\centering
\includegraphics[width=0.47\textwidth]{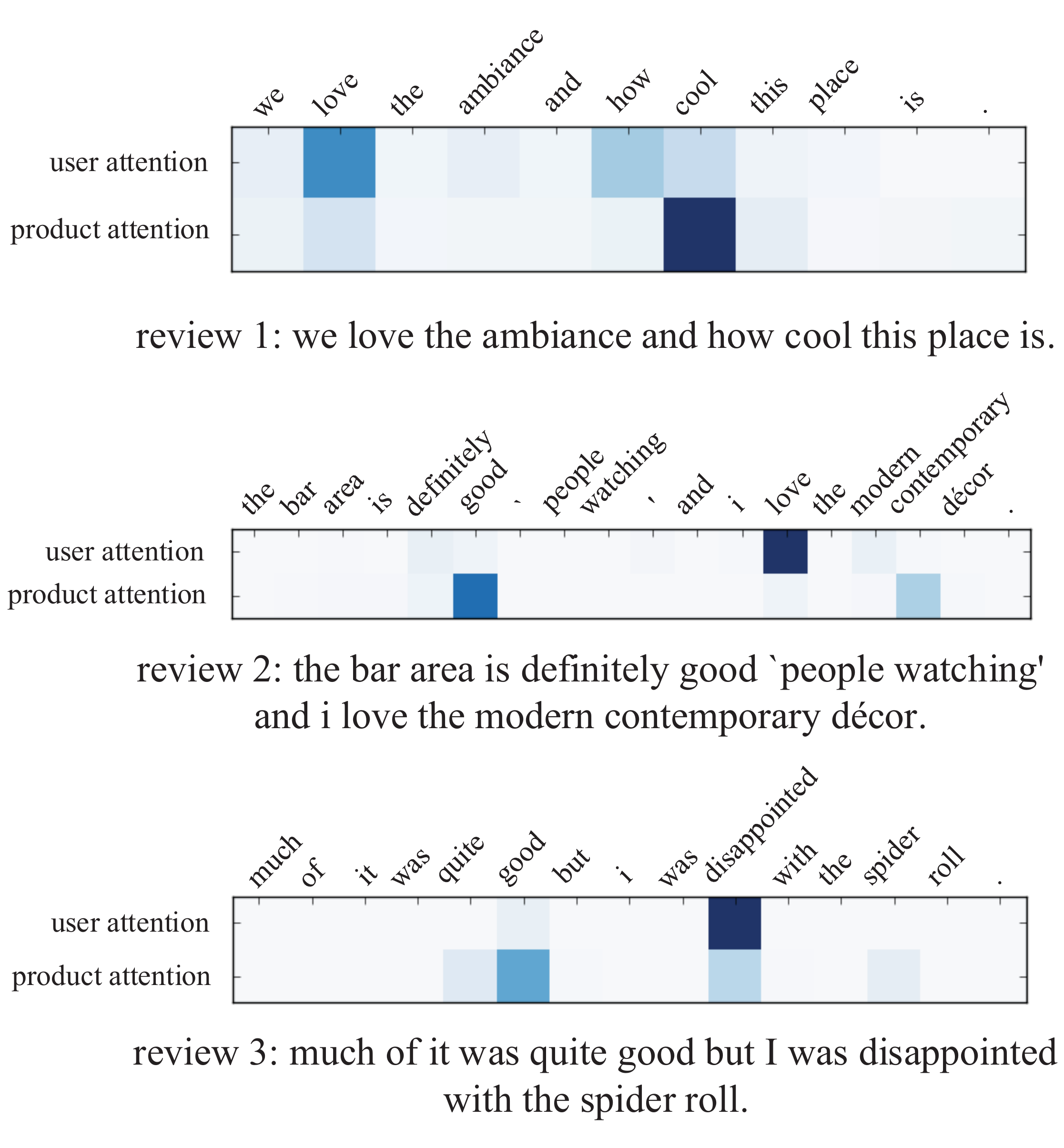}
\caption{Visualization of user attention and product attention over words}
\label{fig:attention}
\end{figure}
To validate our observation and demonstrate the effectiveness of user attention and product attention, we take some review instances in Yelp2013 dataset as example. We visualize the attention weights of these reviews at word level in HUAPA. The results are shown in Figure \ref{fig:attention}. Note that, the darker color means higher weight.

For the review 1, we can see that the word ``\textit{love}" has highest attention weight in user attention and the word ``\textit{cool}" has highest attention weight in product attention. In fact, it is intuitive that ``\textit{love}" usually expresses user's  affection or preference, and ``\textit{cool}" is used to describe the characteristic of product. There are also some reviews that user's preference is inconsistent with product's characteristic, such as review 3 in Figure \ref{fig:attention}. The content of the review 3 is ``\textit{much of it was quite good but I was disappointed with the spider roll.}". It is obvious that the word ``\textit{good}" indicates the product's characteristic and the word ``\textit{disappointed}" shows user's negative sentiment. Our model not only catches such information but also gives right prediction. The visualizations of attention demonstrate that our model can capture global user's preference and product's characteristic.

\section{Conclusion}
In this paper, we propose a novel framework incorporating user and product information for sentiment classification. Specifically, we firstly apply two individual hierarchical neural networks to generate two representations, with user attention or with product attention. Then, we design a combined strategy to make full use of the two representations for training and final prediction. We evaluate our model on several sentiment classification datasets. The experimental results show that our model achieves obvious and consistent improvements compared to other state-of-the-art methods. Finally, the visualizations of attention also show our model can capture user information and product information.

\section{Acknowledgments}
We would like to thank the anonymous reviewers for their insightful comments. This work was supported by the 863 program(No. 2015AA015406) and the NSFC(No. 61472183, 61672277).

\bibliography{huapa}
\bibliographystyle{aaai}
\end{document}